\documentclass[11pt]{article}

\usepackage[preprint]{acl}
\usepackage{tabularx}

\usepackage{times}
\usepackage{latexsym}

\usepackage[T1]{fontenc}
\usepackage[utf8]{inputenc}

\usepackage{microtype}

\usepackage{inconsolata}

\usepackage{graphicx}

\title{Multi-Perspective Transformers in ARC-AGI-2 Challenge}

\author{
Caleb Talley  
\\ {\bf Seun Adekunle} 
\\ {\bf Weiwen Dong}
\\ {\bf Fariha Sheikh} 
\\ {\bf Vedant Tibrewal} 
\\ {\bf Xinyu Wu}
\\ {\tt \{calebtal, sadekunl, weiwendo, fsheikh, vedantti, xwu0427\}@usc.edu}}

\begin{document}
\maketitle
\begin{abstract}
ARC-AGI-2 is a benchmark of human-intuitive visual puzzles that measures a machine’s ability to generalize from limited examples, interpret symbolic meaning, and flexibly apply rules in varying contexts.

In this paper, we discuss our approach to solving the ARC-AGI-2 puzzles with TinyLM, with additional fine-tuning at test time, including Test-Time-Training (TTT) and Products of Experts (POE).

Our model achieves 96.1\% accuracy on the training set and 21.7\% accuracy on the evaluation set.

\end{abstract}

\section{Introduction}
\subsection{Motivation}
The visual puzzles in ARC-AGI-2 are designed to quantify the model's ability to take limited information and apply it to new, unseen instances. A model's intelligence is also determined by the efficiency of its knowledge acquisition—specifically, its ability to translate experience or training data into novel skills in uncertain and highly adaptive environments.

The following Figure \ref{fig:puzzles} displays three basic ARC-AGI puzzles.

\begin{figure}[h!tb]
\includegraphics[width=\linewidth]{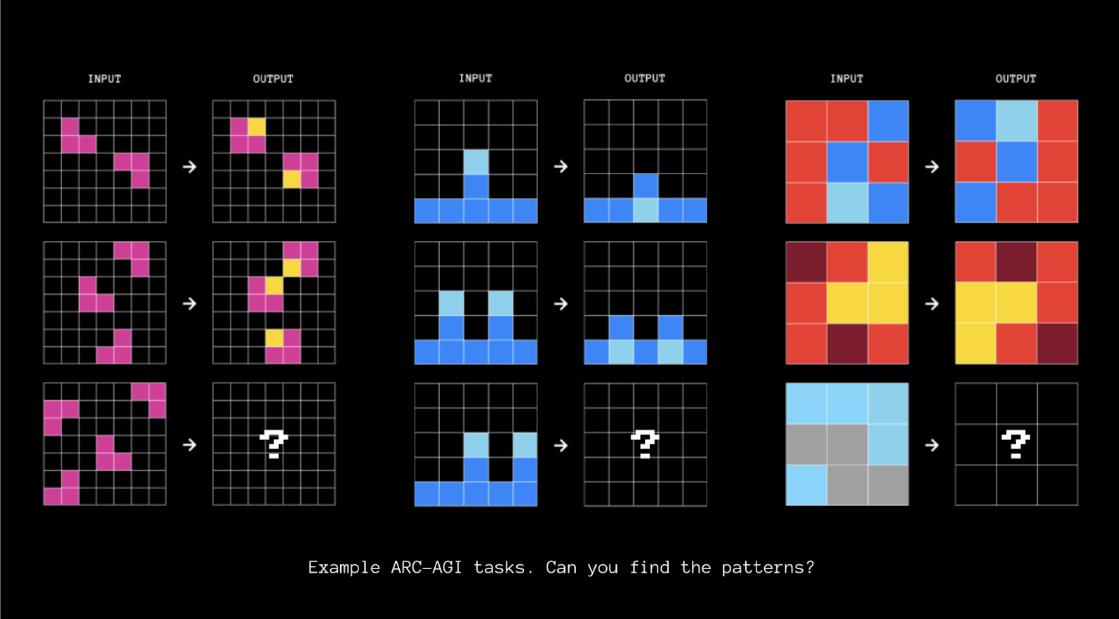}
\caption{Example ARC-AGI 2 Puzzles \cite{new-ideas-for-agi}}
\label{fig:puzzles}
\end{figure}

\subsection{Methods Overview}
The algorithm starts with \textbf{tokenizing} each grid, turning it into a short left-to-right text with small markers for width, height, and colors, so the model understands the grid’s shape and palette. We then create multiple  \textbf{views} of the same puzzle by rotating, flipping, transposing, and relabeling colors. Using the tokenized examples, a compact local  \textbf{generator} proposes candidate outputs for the test grid. We evaluate each candidate with a \textbf{product-of-experts} score, which adds the model’s token-level confidence across all views and keeps only those that also reproduce the given training examples exactly. Finally, we apply light \textbf{test-time training}: a few quick LoRA updates with a minimal learning rate and strong dropout, stopping early when gains stall, so the model adapts to the puzzle’s style without overfitting. For more detailed discussions, refer to the Methods section. 

\section{Related work}
\subsection{Perspective (ARChitects)}

The 2024 ARC-AGI winner works by augmenting grid input to generate varied test samples (views), which are used to fine-tune the NeMo-Minitron-8B model that generates a candidate for each view. Candidates are scored and determined by their ability to stably generate uniform solutions across augmented input views. DFS is used to prune branches with weak probabilities \cite{franzen2025productexpertsllmsboosting}. One drawback of this approach is that it’s less interpretable compared to program synthesis methods.

\subsection{Latent Program Networks (LPNs)}
Programs are represented as vectors in latent space. Using a transformer-based encoder input-output, grid pairs are mapped to a latent program space. At test time, gradient optimization is performed to refine the original program within the latent space. A transformer-based decoder then maps the optimized program and test input to an output grid. As expected, this approach is computationally expensive, and did not fully converge during experimentation \cite{searching-latent-program-spaces}.

\subsection{LLM Transformers with Test Time Training (TTT)}
The approach works by first training the model with 8 billion parameters on numerous generated ARC-like tasks, then at test time, the model is fine-tuned on the specific puzzle’s training examples. This method got second place in the 2024 ARC Prize Pub with 47.5\% accuracy on the semi-private eval dataset \cite{akyurek2024surprising}. 

\subsection{Direct Preference Optimization (DPO)}
This training method teaches models by showing them pairs of “good” and “bad” solution attempts, rather than just correct answers. The model learns to increase the chance of generating preferred solutions while decreasing the chance of generating rejected ones. DPO is simpler than traditional reinforcement learning methods because it skips the step of building a separate reward model \cite{rafailov2023direct}. For ARC-AGI, this could work by generating multiple solution attempts for training puzzles, then having humans or another model judge which attempts are better. The model then learns the patterns that are more likely to generate correct solutions. However, DPO can quickly overfit on small datasets, which is a concern for ARC given its limited training data. The method works best when combined with other techniques rather than used alone.

\section{Methods}
\subsection{Data}
The data used is directly from the ARC-AGI-2 competition, hosted on Kaggle and available on GitHub. ARC-AGI-2 contains 1,000 public training tasks and 120 public evaluation tasks. The semi-private and private evaluation datasets were not available for testing at the time of project completion.
\subsection{Preprocessing}
We have implemented a pipeline to preprocess the
ARC puzzle boards, including grid representation,
view transformation, and serialization.
\subsubsection{Grid}
The Grid type is the fundamental data structure we
use to represent ARC puzzle boards. It establishes
a single source of truth for puzzle representation
across all system components. This prevents silent
bugs that could arise from invalid shapes, accidental data corruption during transformations, 
or inconsistent representations.

It is implemented as a 2-D NumPy array, where
each element is a color identifier. When converted
to a Grid, the input puzzle is strictly validated
against the ARC puzzle specs.
Additionally, we have implemented critical
helper functions, including palette extraction, difference mask generation, and shape validation, to
ensure safe and predictable interactions with the puzzle
data throughout the pipeline.

\subsubsection{Views}
Views are reversible transformations applied to
ARC puzzles to expose underlying patterns that
may be obscured in the original representation.

\subsubsection{Serialization}
Serialization is the process of converting grid data
into a sequence of tokens that large language models can understand. Both encoding and decoding
are implemented: grids can be converted into token
sequences, and token sequences can be converted
back into grids.

\begin{figure}[h!tb] % [h!] attempts to place the figure "here"
    \centering % Centers the figure horizontally
    \includegraphics[width=\linewidth]{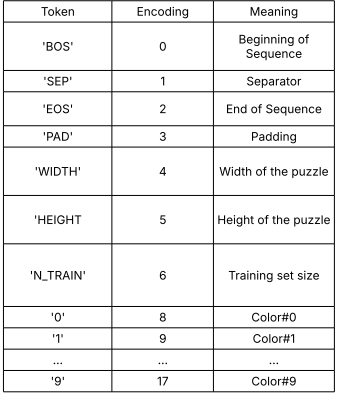} % Replace with your image filename
    \caption{Vocabulary Dictionary} % Your caption text
    \label{fig:dictionary} % A label for cross-referencing
\end{figure}

The serialization process starts with building a
dictionary that maps special tokens (e.g., [BOS]
as the beginning of the sequence) and color identifiers to
a unique number, shown in Figure \ref{fig:dictionary}

\begin{figure}[h!tb] 
    \centering % Centers the figure horizontally
    \includegraphics[width=\linewidth]{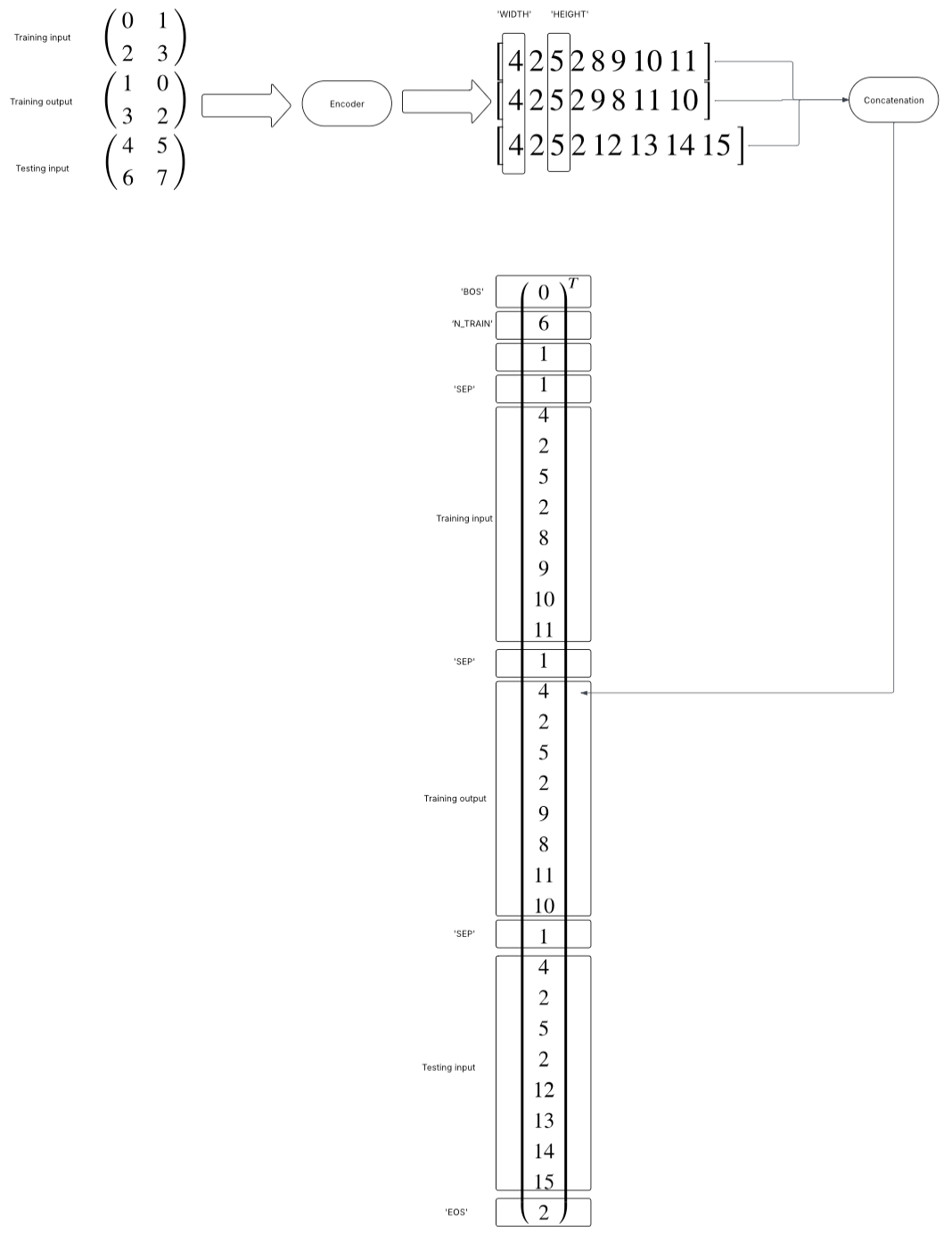} % Replace with your image filename
    \caption{Encoding Algorithm} % Your caption text
    \label{fig:encoding} % A label for cross-referencing
\end{figure}

Then the input grid is encoded as illustrated in
Figure \ref{fig:encoding}.

\subsection{Training}
\subsubsection{Data Augmentation}

We implemented two main types of augmentations:
\begin{itemize}
\item \textbf{Geometric Augmentations} including rotations and reflections
\item \textbf{Color Permutations} such as bijective relabeling of color identifiers
\end{itemize}

This view-augmented approach addresses ARC’s core challenge: puzzles often exhibit solutions that are trivial under certain geometric or color-space transformations, but appear complex in their original presentation. By generating multiple equivalent representations (typically 40-80 views per task), the system increases the probability of finding interpretable patterns that
align with model capabilities.

\subsection{TinyLM Transformer Model}
Our system is built around TinyLM, a decoder-only transformer architecture meticulously optimized for the Abstraction and Reasoning Corpus (ARC) domain. The design centers on balancing representational power, computational efficiency, and generalization capability, prioritizing compactness with a default configuration of approximately 20 million parameters. This choice acknowledges that for the limited ARC training data, excessive model capacity can impose high computational costs during crucial test-time adaptation.

TinyLM adopts the standard transformer decoder structure with several key domain-specific adaptations:
\begin{enumerate}
    \item Positional Embeddings
    
    Utilizes learnable positional embeddings instead of sinusoidal ones to better capture task-specific spatial patterns.
    \item Attention

    Employs causal self-attention, preserving autoregressive generation capabilities.
    \item Activation

Uses GELU activation functions in the feedforward networks for improved gradient flow.
    \item Multi-Head Attention

    Parameterized with 8 heads operating on 56-dimensional subspaces (within a 448-dimensional embedding space). This enables simultaneous processing of multiple relational patterns, essential for coordinating the spatial, chromatic, and structural features in ARC tasks.
    \item Architecture and Stability

    A pre-norm architecture in which layer normalization precedes attention and feedforward sublayers stabilizes training, allowing for a deeper network.
    \item Depth and Connections

    The model consists of an 8-layer stack, where residual connections facilitate robust gradient propagation.
    \item Regularization

A Dropout regularization at a 10\% rate is applied to mitigate overfitting to the limited training data.
    \item Context Capacity

    Supports context windows up to 2048 tokens, ample space for the largest ARC grids (30x30), plus multiple input-output examples needed for test-time training.
\end{enumerate}

This architectural configuration is the result of extensive hyperparameter tuning, successfully balancing validation performance, training stability, and inference efficiency. The final design achieves competitive performance while remaining computationally tractable for the ensemble and adaptation techniques detailed later in the report.

\subsection{Post Training}
\subsubsection{Product of Experts (PoE) Ensemble}
The PoE framework is a robust ensemble prediction method that enhances reliability and mitigates model hallucinations by exploiting the transformation invariance of the multiple views of the same input grid.

\subsubsection{Test-Time Training (TTT)}

TTT is a meta-learning approach that addresses the extreme task diversity in the ARC corpus by adapting a model to each specific task during inference. Since the training data do not cover all unique transformation rules, TTT learns a generic prior during pre-training and then rapidly specializes using the few provided task-specific training examples (typically 2-4 pairs). Inspired by meta-learning frameworks like MAML, TTT involves constructing a small, augmented dataset from the test task's examples, performing a few steps of gradient descent (10-50) with a conservative learning rate to specialize the pre-trained weights, and then predicting the test examples. TTT's effectiveness relies on the quality of test data and hyperparameter configuration. When combined with PoE ensemble prediction, TTT aims to offer improved performance through both task specialization and multi-view robustness.

\section{Results}

We had two evaluation pipelines and used them to evaluate our model as well as Qwen3-8B on the training/testing data The results are shown in Table \ref{tab:model_performance} and Table \ref{tab:test_data_performance}.

\begin{table*}[h]
    \centering
    \caption{Model Performance on Training Set}
    \label{tab:model_performance}
    \begin{tabular}{|l|c|c|c|c|c|}
        \hline
        \textbf{Metrics} & \textbf{Qwen3-8B} & \multicolumn{4}{|c|}{\textbf{Our Model}} \\
        \cline{3-6}
        & \textbf{(Zero-shot)} & \textbf{Pre-trained Only} & \textbf{TTT} & \textbf{PoE} & \textbf{TTT + PoE} \\
        Accuracy & 6.24\% & 96.1\% & 1.9\% & 23.8\% & 30.6\% \\
        \hline
        Valid Generation Rate & 99.4\% & 96.1\% & 1.9\% & 23.8\% & 30.6\% \\
        \hline
        Failed Generation Rate & 0.6\% & 3.9\% & 98.1\% & 76.2\% & 69.4\% \\
        \hline
    \end{tabular}
\end{table*}

\begin{table*}[h]
    \centering
    \caption{Model Performance Comparison on Evaluation Set}
    \label{tab:test_data_performance}
    \begin{tabular}{|l|c|c|c|c|c|}
        \hline
        \textbf{Metrics} & \textbf{Qwen3-8B} & \multicolumn{4}{|c|}{\textbf{Our Model}} \\
        \cline{3-6}
        & \textbf{(Zero-shot)} & \textbf{Pre-trained Only} & \textbf{TTT} & \textbf{PoE} & \textbf{TTT + PoE} \\
        \hline
        Accuracy & 0\% & 21.7\% & 0\% & 0\% & 0\% \\
        \hline
        Valid Generation Rate & 96\% & 21.7\% & 0\% & 0\% & 0\% \\
        \hline
        Failed Generation Rate & 4\% & 78.3\% & 100\% & 100\% & 100\% \\
        \hline
    \end{tabular}
\end{table*}

We evaluated several strategies (Product-of-Experts, Test-Time training) in various combinations (TTT, PoE, TTT + PoE). We go into more detail below:

\subsection{Pipeline 1}

\begin{figure}
    \centering
    \includegraphics[width=1\linewidth]{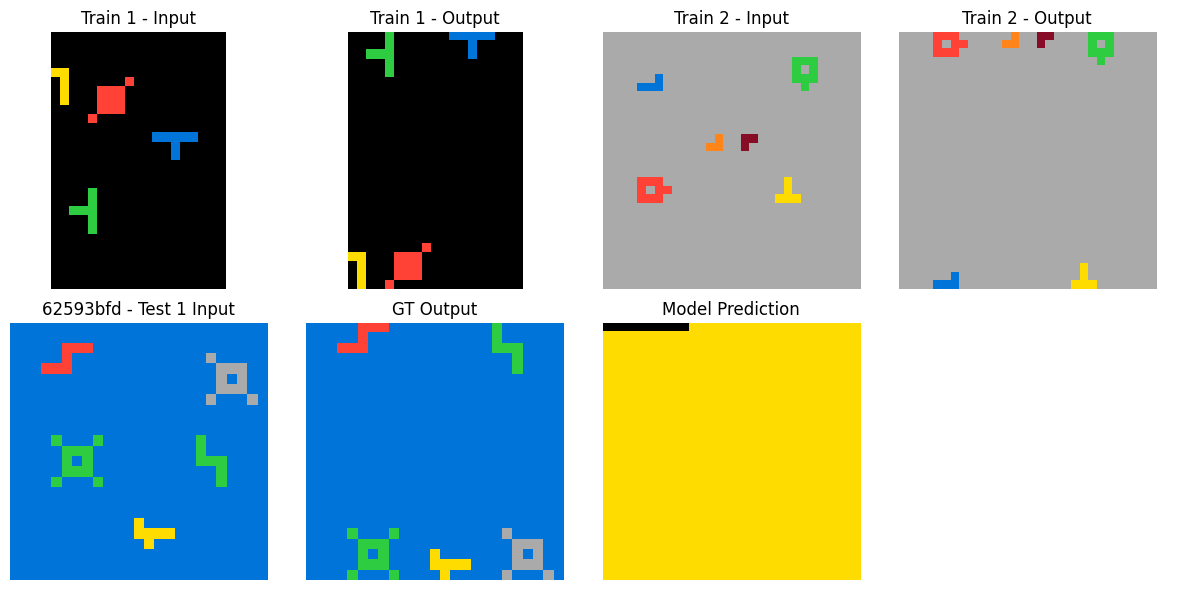}
    \caption{Model evaluation before few-shot prompting}
    \label{fig:model-before}
\end{figure}

\begin{figure}
    \centering
    \includegraphics[width=1\linewidth]{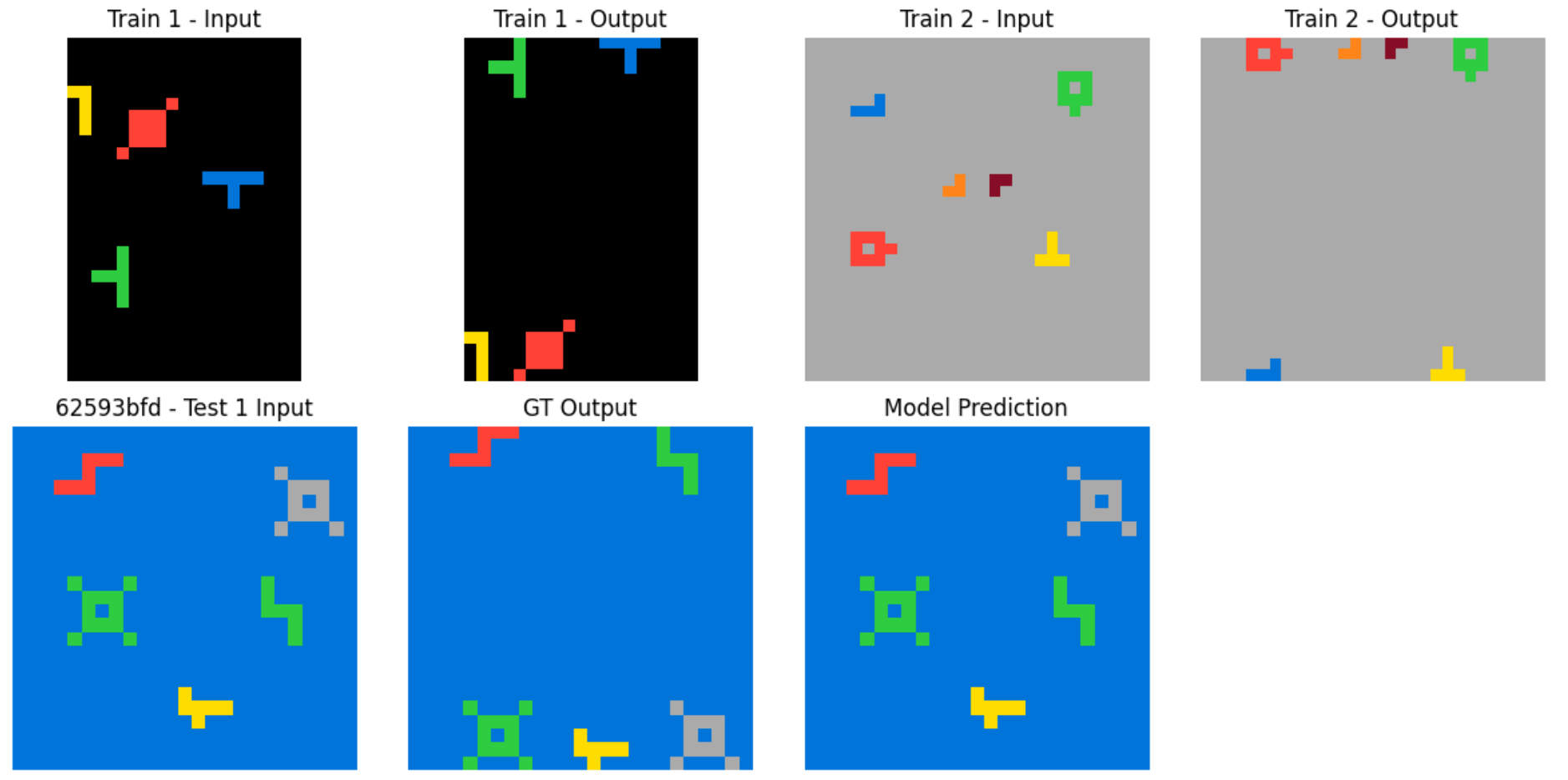}
    \caption{Model evaluation after few-shot prompting}
    \label{fig:model-after}
\end{figure}

Given the trained checkpoint, we load a TinyLM model and iterate through each task. For each task, we fine-tune the model using its training examples to maximize performance on that task individually. We also integrated multiple orientations to find the best “view” where the model understands the pattern for a specific test best. After applying the best view, we proceeded to integrate few-shot prompting, which involved appending three training examples directly to the context. Few-shot prompting was able to improve the model's inference regarding task rules and enable in-context meta-learning. As shown in Figure \ref{fig:model-before} \& \ref{fig:model-after}, the model initially resets to a default generic pattern but later improves to understand how each color/shape is relocated, leading to a more coherent output and showing improvement in intermediate reasoning. However, on the evaluation tasks, the results remained the same as those of the second pipeline, with no further gains achieved.

\subsection{Pipeline 2}
The second pipeline was designed as a systematic ablation study to isolate the contributions of individual inference strategies. We implemented a multi-strategy evaluator that tests four configurations: Baseline (greedy generation from TinyLM), PoE, TTT, and their combination (TTT+PoE). Each strategy was evaluated independently on both the training set (1,000 tasks) and the evaluation set (120 tasks). 

For PoE, we attempted to combine multiple serialization "views" (row-major, column-major) under the hypothesis that different orderings capture complementary spatial patterns. However, we discovered that our model was trained on row-major serialization, rendering alternative views ineffective. 

For TTT, we fine-tuned the model on each task’s training examples before inference (10 steps, lr=5e-5). We hypothesized that task-specific adaptation would improve test performance. Surprisingly, TTT often degraded performance, particularly on the evaluation set where it achieved near 0\% accuracy. The model appeared to overfit to training example idiosyncrasies that did not transfer to test inputs.

The combined TTT+PoE strategy inherited both limitations, resulting in the lowest overall performance. Our TinyLM achieved 96.1\% accuracy on training tasks and 21.7\% on evaluation tasks. PoE was limited by the fact that the model was trained, for the most part, on row-major serialization. TTT exhibited frequently degraded performance due to overfitting, achieving near 0\% on evaluation tasks. The substantial training-evaluation gap (74.4\%) indicates that while our model masters learned transformations, generalizing to novel patterns requires improvements in both architectural and training methodologies rather than relying solely on inference-time techniques.

\section{Discussion on Future Work}
Due to time and resource constraints, there are areas of improvement that could be applied to our methodology.

\subsection{Training Data Expansion} The model’s limited performance on evaluation tasks suggests it needs exposure to more diverse patterns. Adding the full ARC-AGI 1 dataset and using Re-ARC could help the model learn more general transformation rules rather than memorizing specific examples.

\subsection{Test-Time Training Improvements} TTT degraded performance because it overfitted to the few training examples in each task. Future work should explore regularization techniques during TTT to prevent this, such as dropout or weight decay. Using more fine-tuning steps with a smaller learning rate (lower than our current 5e-5) could allow gentler adaptation.

\subsection{Product-of-Experts Refinement} PoE failed because our model was primarily trained on row-major serialization, making other views useless. To address this, we should train the model on multiple serialization formats from the start (e.g., row-major, column-major, and others) so that different views capture genuinely distinct patterns.

\subsection{Model Scaling} Moving from TinyLM to a small-sized model could improve generalization simply through increased capacity. Larger models, with, for example, 50M-100M parameters, typically handle novel patterns better, though this comes with higher computational costs.

\subsection{Decoding and Generation} Our strict decoding rules may reject valid solutions due to minor formatting issues. Implementing more flexible output validation that focuses on grid correctness rather than exact token sequences could recover lost accuracy. We could also explore beam search or sampling strategies as alternatives to greedy decoding.

\subsection{Interpretability} The model overgenerates outputs for larger puzzles (over 20x20), causing shape mismatches and default failures. Applying interpretability techniques can help analyze these patterns and develop strategies to reduce overgeneration for large puzzles.


\bibliography{custom}
\section{Appendix}

\subsection{Code}

\begin{itemize}
  \item Project: \href{https://github.com/CalebTalley2024/ARC-AGI-2}{https://github.com/CalebTalley2024/ARC-AGI-2}
  \item Dataset: \href{https://github.com/arcprize/ARC-AGI-2}{https://github.com/arcprize/ARC-AGI-2}
  \item Benchmarking: \href{https://github.com/arcprize/arc-agi-benchmarking}{https://github.com/arcprize/arc-agi-benchmarking}
\end{itemize}

\end{document}